\documentclass{article}
\usepackage[preprint]{spconf}
\copyrightnotice{\begin{tabular}[t]{@{}l@{}}\copyright 2020 IEEE. Personal use of this material is permitted. Permission from IEEE must be obtained for all other uses, in any cur- \\rent or future media, including reprinting/republishing this material for advertising or promotional purposes, creating new col-\\lective works, for resale or redistribution to servers or lists, or reuse of any copyrighted component of this work in other works.\end{tabular}}

\usepackage{spconf,amsmath,graphicx, float, caption, threeparttable, amssymb, array, enumitem}
\usepackage[utf8]{inputenc}

\title{DESHUFFLEGAN: A Self-Supervised GAN to improve structure learning}
\name{Gulcin Baykal, Gozde Unal}
\address{Istanbul Technical University, 
        Department of Computer Engineering, 
        Istanbul, Turkey}
\begin{document}
%

\maketitle
\begin{abstract}
Generative Adversarial Networks (GANs) triggered an increased interest in problem of image generation due to their improved output image quality and versatility for expansion towards new methods. Numerous GAN-based works attempt to improve generation by architectural and loss-based extensions. We argue that one of the crucial points to improve the GAN performance in terms of realism and similarity to the original data distribution is to be able to provide the model with a capability to learn the spatial structure in data. To that end, we propose the  $\textit{DeshuffleGAN}$ to enhance the learning of the discriminator and the generator, via a self-supervision approach. Specifically, we introduce a deshuffling task that solves a puzzle of  randomly shuffled image tiles, which in turn helps the DeshuffleGAN learn to increase its expressive capacity for spatial structure and realistic appearance. We provide experimental evidence for the performance improvement in generated images, compared to the baseline methods, which is consistently observed over two different datasets.
\end{abstract}
\section{INTRODUCTION}

\begin{figure}[t]
    \centering
    \includegraphics[width=0.98\linewidth]{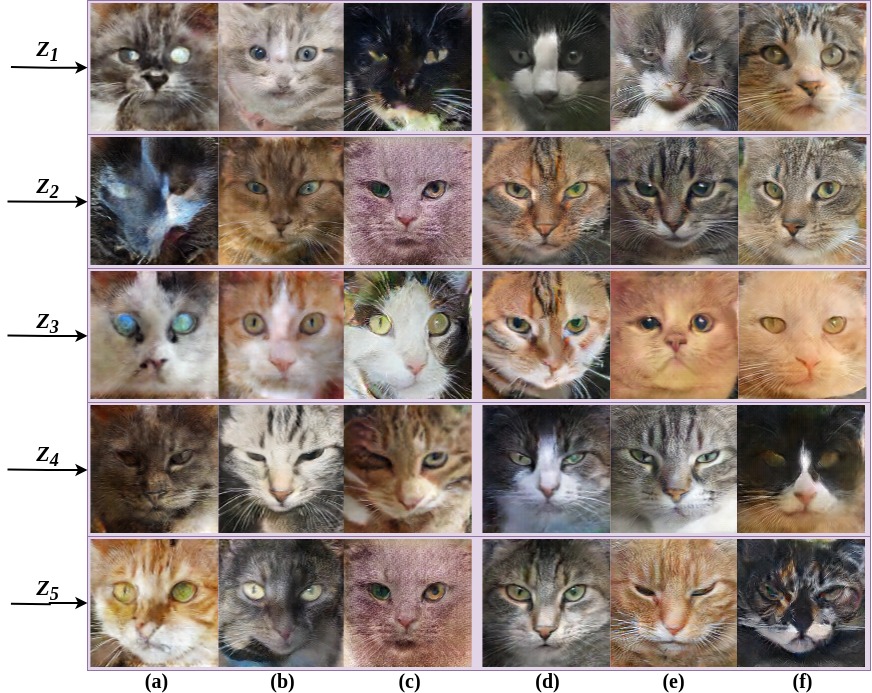}
    \caption{\small Generation results for CAT dataset. 5 different vectors are sampled from the normal distribution, and are given as input to 6 different GAN models: (a) RaSGAN, (b) RaLSGAN, (c) RaHingeGAN, (d) Deshuffle(RaS)GAN, (e) Deshuffle(RaLS)GAN, (f) Deshuffle(RaHinge)GAN. Compared to the base models in (a), (b) and (c), DeshuffleGANs in (d), (e) and (f) are observed to generate more structured and realistic results.}
    \label{fig:cats_inference}
\end{figure}

In unsupervised learning, Generative Adversarial Networks (GANs) \cite{gan} framework enables deep neural network models to generate data samples with distributions similar to a desired input data distribution. As GANs recently achieved a high level of performance in terms of photo-realism, particularly during the last few years, numerous GAN variants are constructed for different tasks involving unconditional image generation \cite{progan, stylegan} and conditional image generation \cite{cgan, acgan, biggan}. Improving a GAN's generation quality relies on the GAN model to be able to learn the underlying structure in data. In unconditional generation, quality of the generations are increased by models with massive number of parameters \cite{progan, stylegan} which are computationally expensive to use. Generator part of the GAN, which is essentially a decoder network that generates new samples, should be supported by different guidance strategies to increase its expressiveness, ideally in a way that it does not require massive networks. To that end, in this paper, we augment the GAN model with a new component that is relatively light in terms of computation, and that enriches its capture of the structural properties in the images in terms of underlying spatial order of image tiles. 

\begin{figure*}[htb]
    \centering
    \includegraphics[width=0.99\linewidth]{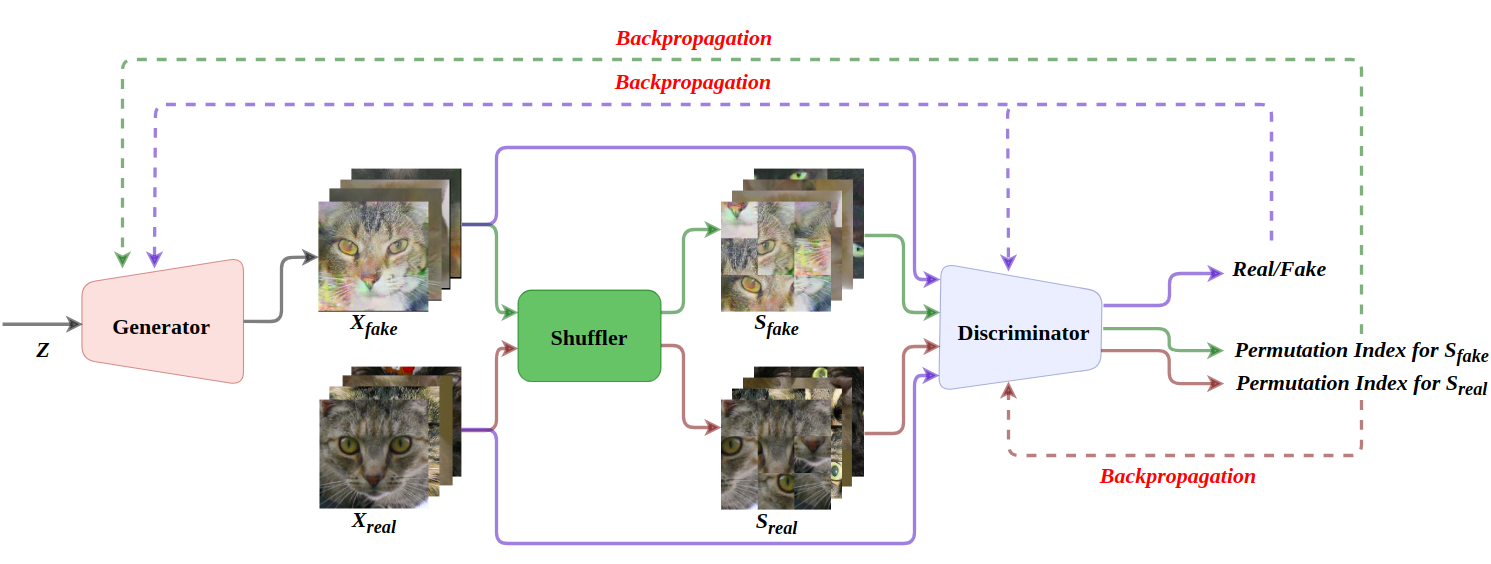}
    \caption{\small Architecture of DeshuffleGAN. Green and Red lines indicate the shuffling and deshuffling operations for $X_{fake}$ and $X_{real}$, respectively, including their backpropagation paths. Purple lines refer to the classical adversarial training of the original GAN.}
    \label{fig:main_figure}
\end{figure*}


We introduce a new model, the \textit{DeshuffleGAN}, where we construct a discriminator whose task is to deshuffle a shuffled input into its correct tile order while also carrying out its original task. We show that it both increases the performance of the discriminator to learn the input data structure and improves the quality of generator due to structurally relevant feedback provided by the deshuffling process. Sample generation results for DeshuffleGAN on CAT dataset \cite{cats} are shown in Figure \ref{fig:cats_inference}. The results support our hypothesis in that additional task of predicting a shuffling order of shuffled input data for the discriminator on top of classical adversarial training task which only focuses on classification of the input data as real and fake, improves the generation quality, which is also demonstrated by our quantitative performance results. 

Contributions of our approach are summarized as follows:

\begin{itemize} [noitemsep, nolistsep]
    \item We design a discriminator with shared weights for two discriminative tasks: real/fake (r/f) classification and deshuffling. 
    \item We define a new deshuffling loss to improve the performance of unconditional generation. This loss feeds the generator an error signal to aid it in producing structurally photorealistic images. A deshuffler discriminator is not used before in Self-Supervised GANs. 
    \item We achieve better performance results in terms of Fréchet Inception Distance (FID) \cite{fid} measure for the DeshuffleGAN compared to the baseline methods.
\end{itemize}

\section{RELATED WORK}

\textbf{Self-supervised learning} is a variant of unsupervised learning that pushes the classifier models, that recently take mostly the form of deep neural networks, to learn a better encoding of the input by providing the model an additional task to solve. In self-supervised learning, this task is called \textit{pretext} and the labels of the data to solve the pretext are pseudo-labels, as they are obtained by intentionally modifying the input data in a certain way so that the true output of the prediction is already known. Some widely used pretexts are predicting the rotation angles of the images \cite{rotation}, predicting the permutation order for image tiles randomly shuffled \cite{jigsaw, jigen}, and colorizing a grayscale image \cite{colorization}.

For the mentioned tasks, as the labels of data come free, pretext task can be solved in a supervised manner. Then, the learned feature representations can be beneficial to solve other tasks by transferring those features to other network models.

\textbf{Generative models}, on the other hand, received an increased attention after the two adversarial player idea of GANs is introduced in \cite{gan}. In \cite{dcgan}, usage of convolutions in GANs is shown and different GAN variants are constructed within the last years to help stabilize their training, which is known to be not robust. Some GAN methods attempt to improve the stability of adversarial training \cite{gp, spectral, ragan} while others try to improve the generation quality \cite{stylegan, biggan}. 

Recently, self-supervised learning has been applied to GANs in order to prevent the catastrophic forgetting problem of the discriminator \cite{ssgan}. In \cite{ssgan}, the  self-supervision (SS) task of image rotation prediction of four rotation angles in 2D plane is selected, and rotation-based SS task is shown to help the discriminator to learn more structured feature representations. In \cite{ssgan_minimax}, a new SS task, which is based on a multi-class minimax game is proposed in order to support the GAN model to capture the data distribution better.

In this paper, we employ deshuffling as an SS task as it encodes both global and local structural order in an image while its complexity compared to rotation prediction furthers the discriminator's quest to improve the GAN model's expressiveness.

\section{METHODOLOGY}

Architecture of the proposed DeshuffleGAN is shown in Figure \ref{fig:main_figure} with its training mechanism. Generator $G$ takes a vector \textit{z}, sampled from the normal distribution, and returns $X_{fake}$, the generated samples. $X_{real}$ represents the samples from the original data distribution. Both $X_{real}$ and $X_{fake}$ are shuffled by the \textit{Shuffler}. $S_{fake}$ and $S_{real}$ are obtained as the shuffled version of $X_{fake}$ and $X_{real}$, respectively. 

Shuffler divides its inputs into 
9 square tiles of width and height of \textit{input size / 3}. If the input size is not divisible by 3 which is also the case for this work since the generations are 128x128, 
then the biggest number smaller than the input size and divisible by 3 is chosen. Next, the tiles are shuffled with the order of randomly selected permutation. For shuffling, out of the 9! possible permutations, which is unnecessarily huge, 30 different permutations are selected according to Hamming distances between the permutations as in \cite{jigen}. Each sample in an input batch is shuffled by a different, random permutation out of these 30 permutations. Shuffled samples are padded in order to obtain the same output size with the input size. 

Discriminator $D$ classifies $X_{real}$ and $X_{fake}$ as r/f as in a standard GAN setting. In DeshuffleGAN, the {\it same} discriminator predicts the permutation indices of shuffling orders for both $S_{real}$ and $S_{fake}$. $D$ shares the same weights for both tasks except for the output layers. DeshuffleGAN discriminator is trained to be able to predict the shuffling orders so that when the generator starts to generate samples that are well-structured and include related, meaningful pieces including continuity in structures, their jigsaw puzzles can be deshuffled by the discriminator. If the generation results are not structured and do not include meaningful pieces, then the puzzle tiles will not be related to each other, and the discriminator will give a negative feedback to the generator to generate structured, meaningful samples.

\subsection{Adversarial Loss}
Objective functions for classical GAN training are given by:
\begin{align}
    L_D &= -\mathbb{E}_{x_r \sim P} [\log(D(x_r))] - \mathbb{E}_{x_f \sim Q} [\log(1 - D(x_f))] \nonumber\\
    L_G &= -\mathbb{E}_{x_f \sim Q} [\log(D(x_f))] \nonumber\\
    D(x_r) &= sigmoid(C(x_r))  \nonumber\\
    D(x_f) &= sigmoid(C(x_f))  \label{eq:1} 
\end{align}
where $x_r$ is the real data, $P$ is the real data distribution, $x_f$ is the generated data, $Q$ is the generated data distribution and $C(x)$ is some measure of realness of input $x$. $L_D$ and $L_G$ refer to loss functions for the $D$ and $G$ networks.

In classical GAN training, $D$ estimates the probability of the input to be real while $G$ tries to increase the probability of fake data to be deemed as real. It is argued in RaGAN (Relativistic Average GAN) \cite{ragan}, that this idea leads to a problem in training because $G$ pushes $D$ to output 1 for both the real and fake data whereas in fact the discriminator should converge to 0.5 to realize JS-divergence between input and generated data distributions. It is stated that the aim of the training should be not only to increase the probability that the fake data is real, but also to decrease the probability that the real data is real. These observations lead to the new objective in RaGAN, which introduces relativism such that the discriminator will estimate the probability of input data being more realistic than the generated data.

As the relativistic objectives improve the quality of generations and enable a more stable and faster training mechanism, we use RaGAN loss-based GANs \cite{ragan} as the baseline method. For the generator and the discriminator, DCGAN \cite{dcgan} architecture is utilized as in \cite{ragan}. For the permutation task only, we add one more convolutional layer to the output of the discriminator. In DeshuffleGAN, the losses for the $D$ (r/f task only part) and the $G$ networks are set as follows:
\begin{align}
    L_D &=  -\mathbb{E}_{x_r \sim P} [\log(\tilde{D}(x_r))] - \mathbb{E}_{x_f \sim Q} [\log(1 - \tilde{D}(x_f))] \nonumber\\
    L_G &= -\mathbb{E}_{x_f \sim Q} [\log(\tilde{D}(x_f))] - \mathbb{E}_{x_r \sim P} [\log(1 - \tilde{D}(x_r))] \nonumber\\
    \tilde{D}(x_r) &= sigmoid(C(x_r) - \mathbb{E}_{x_f \sim Q}C(x_f)) \nonumber\\
    \tilde{D}(x_f) &= sigmoid(C(x_f) - \mathbb{E}_{x_r \sim P}C(x_r)) \label{eq:2} 
\end{align}
$D$ predicts the r/f probabilities both for $X_{real}$ and $X_{fake}$ since the shuffled data do not affect the standard adversarial objective.

\subsection{Deshuffling Loss}
In order to introduce the capability of deshuffling an unordered version of the images, hence increasing the learning capacity of the discriminator and the generator, DeshuffleGAN utilizes a new loss term, called the Deshuffling Loss. 

For the discriminator, the objective is to minimize the error between the true shuffling order and the prediction for shuffling order of $S_{real}$. The reason for updating $D$ only according to $S_{real}$ is: via deshuffling to learn inclusively and solely the features for real data. If $D$ were also to be updated according to $S_{fake}$, $D$ will try to learn the fake and not necessarily meaningful structure of the data, hence the generator will not be able to generate samples compatible to the real data distribution, which is obviously undesirable.

For the generator, the objective is to minimize the error between the true shuffling order  and the prediction for shuffling order of $S_{fake}$. The reason for updating the generator according to $S_{fake}$ is to generate qualified samples that can fool the discriminator. If the generated samples are structured and compatible to the real data distribution, the discriminator that learns the features of the real data can deshuffle them and the discriminator gives a positive feedback to the generator. In other cases when the generated samples are not structured and do not include features compatible to the real data features, then the feedback will be negative.

The deshuffling objectives of the discriminator and the generator are given by the cross-entropy loss as $V_{disc}$ and $V_{gen}$ respectively: 
\begin{equation}
    V_{disc} = -\sum_n^N y_d^n \ln \tilde{y}_d^n 
    \quad\mathrm{,}\quad 
    V_{gen} = -\sum_n^N y_g^n \ln \tilde{y}_g^n
    \label{eq:DeShuffleLoss}
\end{equation}
where $N$ denotes number of samples, $y_d$ is the one-hot encoded label vector of size 30x1 for $S_{real}$, $\tilde{y}_d$ is the prediction vector of the permutation index for $S_{real}$, $y_g$ is the one-hot encoded label vector of size 30x1 for $S_{fake}$, and $\tilde{y}_g$ is the prediction vector of the permutation index for $S_{fake}$.

\subsection{Full Objective}
Full objective for the training of DeshuffleGAN is given by:
\begin{equation}
    L_D = L_D + \alpha V_{disc}
    \quad\mathrm{,}\quad 
    L_G = L_G + \beta V_{gen}
    \label{eq:DeshuffleGANLoss}
\end{equation}
where the parameters $\alpha$ and $\beta$ decide how effective the different objective terms will be in the training. $\alpha$ is selected as 1 and $\beta$ is selected as 0.2 in our experiments as in \cite{ssgan}.

\begin{figure}[t]
    \centering
    \includegraphics[width=\linewidth]{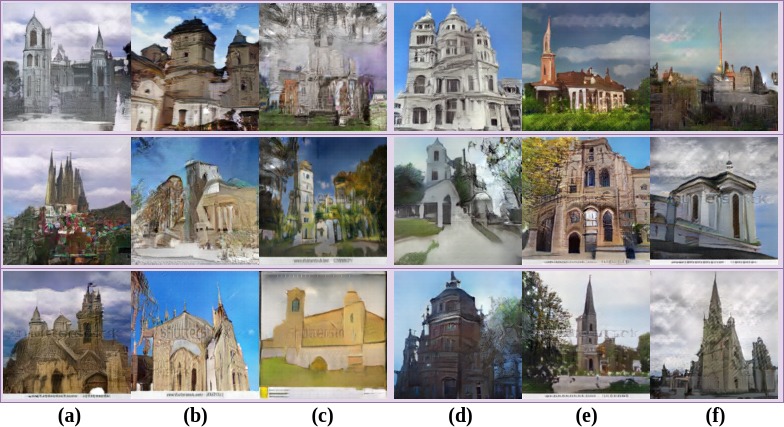}
    \caption{\small Generation results for LSUN Church dataset: (a) RaSGAN, (b) RaLSGAN, (c) RaHingeGAN, (d) Deshuffle(RaS)GAN, (e) Deshuffle(RaLS)GAN, (f) Deshuffle(RaHinge)GAN.}
    \label{fig:lsun_inference}
\end{figure}

\section{EXPERIMENTS}

\textbf{Datasets:} DeshuffleGAN and the baseline GANs are evaluated for CAT \cite{cats} and LSUN Church \cite{lsun} datasets. CAT dataset consists of 5.652 samples with the size of 128x128 or bigger, and comes as preprocessed in order to capture only the faces of the cats. That is why we start with CAT dataset whose distribution has a certain expected structure, namely cat faces. Similarly, we experiment with LSUN Church dataset, which includes buildings that also carry a certain structure. LSUN Church dataset is more challenging than the CAT dataset since its diversity is wider and it consists of 126.227 samples. To our knowledge, this is the first work that shows the evaluation results of the to-be mentioned GAN-baseline methods on LSUN Church dataset. 

\textbf{Objective Functions:} RaSGAN with the standard adversarial training loss in \cite{gan}, RaLSGAN with the least squares loss  \cite{lsgan}, and RaHingeGAN with the hinge loss \cite{geogan} are used as baseline methods. DeshuffleGAN versions of the baselines add the deshuffling losses as in Eq.~(\ref{eq:DeshuffleGANLoss}).

\textbf{Experimental Settings and Hyperparameters:} Experiments are conducted on an NVIDIA Quadro P6000 GPU and PyTorch framework is used for the implementations. Adam \cite{adam} is used as the optimizer for both of the networks. Batch size is selected as 32. For a fair comparison with the baseline methods, the random seed is selected as 1 in all of the experiments. The number of total iterations is selected as 100K for CAT dataset, 300K for LSUN Church dataset. 

\begin{table}[t]
  \begin{center}
    \caption{\small Evaluation results on CAT and LSUN Church datasets ($\textbf{*}$: Minimum FID values reported in \cite{ragan})}
    \label{tab:eval}
    \begin{tabular}{|c|c|c|c|c|} 
        \hline
      \textbf{Network} & \multicolumn{2}{c|}{\textbf{LSUN Church}}  & \multicolumn{2}{c|}{\textbf{CAT}}\\
      \cline{2-5}
      & FID$_1$ & FID$_2$ & FID$_1$ & FID$_2$\\ 
      \hline
      RaSGAN & 29.02 & 29.60 & 21.05\textbf{*} & 62.89 \\
      \hline
      Deshuffle & & & & \\
      (RaS)GAN & 28.41 & 28.89 & 21.91 & 25.09\\
      \hline
      RaLSGAN & 21.05 & 26.24 & 15.85\textbf{*} & 33.75\\
      \hline
      \textbf{Deshuffle} & & & & \\
      \textbf{(RaLS)GAN} & \textbf{20.23} & 21.74 & \textbf{15.10} & 20.90 \\
      \hline
      RaHingeGAN & 24.77& 41.55 & 22.07 & 41.77 \\
      \hline
      Deshuffle & & & & \\
      (RaHinge)GAN & 23.55 & 24.43 & 21.18 & 28.60\\
      \hline
    \end{tabular}
  \end{center}
\end{table}

Table \ref{tab:eval} shows the evaluation results on both datasets in terms of FID. FID \cite{fid} is the most commonly used evaluation metric to measure the generation quality. Lower FID means the data distributions of real samples and generated samples are more similar as desired. FID values of the best-performing models for both datasets are reported in FID$_1$ columns. Additionally, FID values for the generators that have the minimum $L_G$ values are reported in FID$_2$ columns. Differences between the values in FID$_1$ and FID$_2$ show that the generation quality is not correlated with the loss values. DeshuffleGANs achieve lower FIDs with respect to the baselines in all of the settings except RaSGAN on CAT dataset.

\textbf{Deshuffle(RaLS)GAN} achieves the highest performance on both datasets. Visual results are shown in Figure \ref{fig:cats_inference} for CAT dataset and in Figure \ref{fig:lsun_inference} for LSUN Church dataset, where 3 different random vectors are the inputs. It can be observed that the results of DeshuffleGANs depict higher visual quality with more structured and realistic looking images compared to those of the baselines. Generation results validate our intuition that the deshuffling task helps the discriminator to learn the structures in the data.
    
\section{CONCLUSION AND FUTURE WORK}

In this paper, we present the DeshuffleGAN, where a new SS task for the discriminator is introduced to increase its representation power through improved structure learning. With addition of self-supervised task of deshuffling the shuffled input samples, we demonstrate both quantitatively in terms of FID measure and visually that DeshuffleGANs consistently improve the generation quality compared to the baseline methods based on very recent relativistic average GANs. As future work, the deshuffling task can be assigned to different GAN discriminators in order to further test its utility. 


\section{ACKNOWLEDGEMENTS}
In this work, Gulcin Baykal was supported by the Turkcell-Istanbul Technical University  Researcher Funding Program. We gratefully acknowledge the support of NVIDIA Corporation with the donation of the Quadro P6000 used for this research.



\bibliographystyle{IEEEbib}
\bibliography{refs}

\end{document}